\documentclass[lettersize,journal]{IEEEtran}
\usepackage{amsmath,amsfonts}
\usepackage{algorithmic}
\usepackage{array}
\usepackage{textcomp}
\usepackage{stfloats}
\usepackage{url}
\usepackage{color} 
\usepackage{diagbox}
\usepackage{hyperref}
\usepackage{makecell}
\usepackage{subfigure} 
\usepackage{verbatim}
\usepackage{graphicx}
\usepackage{tabularx}
\usepackage{booktabs}
\usepackage{orcidlink} 
\hypersetup{hidelinks} 
\def\BibTeX{{\rm B\kern-.05em{\sc i\kern-.025em b}\kern-.08em
    T\kern-.1667em\lower.7ex\hbox{E}\kern-.125emX}}
\usepackage{balance}
\begin{document}
\title{Global Semantic-Guided Sub-image Feature Weight Allocation in High-Resolution Large Vision-Language Models}

\author{Yuxuan Liang$^{\orcidlink{0009-0004-5039-7568}}$, Xu Li$^{\orcidlink{0009-0001-2431-3410}}$, Xiaolei Chen$^{\orcidlink{0009-0005-5700-9326}}$, Haotian Chen$^{\orcidlink{0009-0001-0593-5281}}$, Yi Zheng$^{\orcidlink{0009-0006-1549-6979}}$, \\ Chenghang Lai$^{\orcidlink{0000-0001-9181-9334}}$, Bin Li$^{\orcidlink{0000-0002-9633-0033}}$, Xiangyang Xue\textsuperscript{$\ast$}$^{\orcidlink{0000-0002-4897-9209}}$, Member, IEEE
\thanks{Manuscript received xx xx, xx; revised xx xx, xx. This work is supported by Shanghai Key Laboratory of Intelligent Information Processing and School of Computer Science, Fudan University. (Corresponding author: Xiangyang Xue.)}


\thanks{Y Liang, X Li, X. Chen, H. Chen, Y Zheng, C Lai, B Li and X. Xue are with the School of Computer Science, Fudan University, Shanghai 200433, China (E-mails: {yxliang23, xu\_li23, chenxl23, htchen24, yizhengplus@gmail.com, chlai21}@m.fudan.edu.cn, libin@fudan.edu.cn, xyxue@fudan.edu.cn)}

}

\markboth{Journal of \LaTeX\ Class Files,~Vol.~xx, No.~x, September~202x}%
{How to Use the IEEEtran \LaTeX \ Templates}

\maketitle


\begin{abstract}
As the demand for high-resolution image processing in Large Vision-Language Models (LVLMs) grows, sub-image partitioning has become a popular approach for mitigating visual information loss associated with fixed-resolution processing. However, existing partitioning methods uniformly process sub-images, resulting in suboptimal image understanding. In this work, we reveal that the sub-images with higher semantic relevance to the entire image encapsulate richer visual information for preserving the model's visual understanding ability. Therefore, we propose the Global Semantic-guided Weight Allocator (GSWA) module, which dynamically allocates weights to sub-images based on their relative information density, emulating human visual attention mechanisms. This approach enables the model to focus on more informative regions, overcoming the limitations of uniform treatment. We integrate GSWA into the InternVL2-2B framework to create SleighVL, a lightweight yet high-performing model. Extensive experiments demonstrate that SleighVL outperforms models with comparable parameters and remains competitive with larger models. Our work provides a promising direction for more efficient and contextually aware high-resolution image processing in LVLMs, advancing multimodal system development.
\end{abstract}


\section{Introduction}
\IEEEPARstart{I}{n}  recent years, Large Vision-Language Models (LVLMs) have made significant strides, advancing multimodal understanding across a range of vision-language tasks. The development of various LVLMs, such as LLaVA \cite{liu2024llava}, InternVL \cite{chen2024internvl}, and ShareGPT4V \cite{chen2025sharegpt4v}, has substantially broadened the scope and potential of cross-modal applications involving both vision and language.  With the increasing demand for more detailed and accurate visual comprehension, high-resolution image processing has become increasingly critical for LVLMs, as such images contain more fine-grained information \cite{monkey}. However, traditional LVLM architectures often face limitations in their ability to effectively process high-resolution images, primarily due to the constraints of their vision encoders, which are typically trained to handle lower-resolution inputs \cite{dong2024internlmxcomposer4khd,monkey}.

\begin{figure}[t]
    \centering
    \subfigure[Sleigh in Polar]{
        \includegraphics[width=0.4\columnwidth]{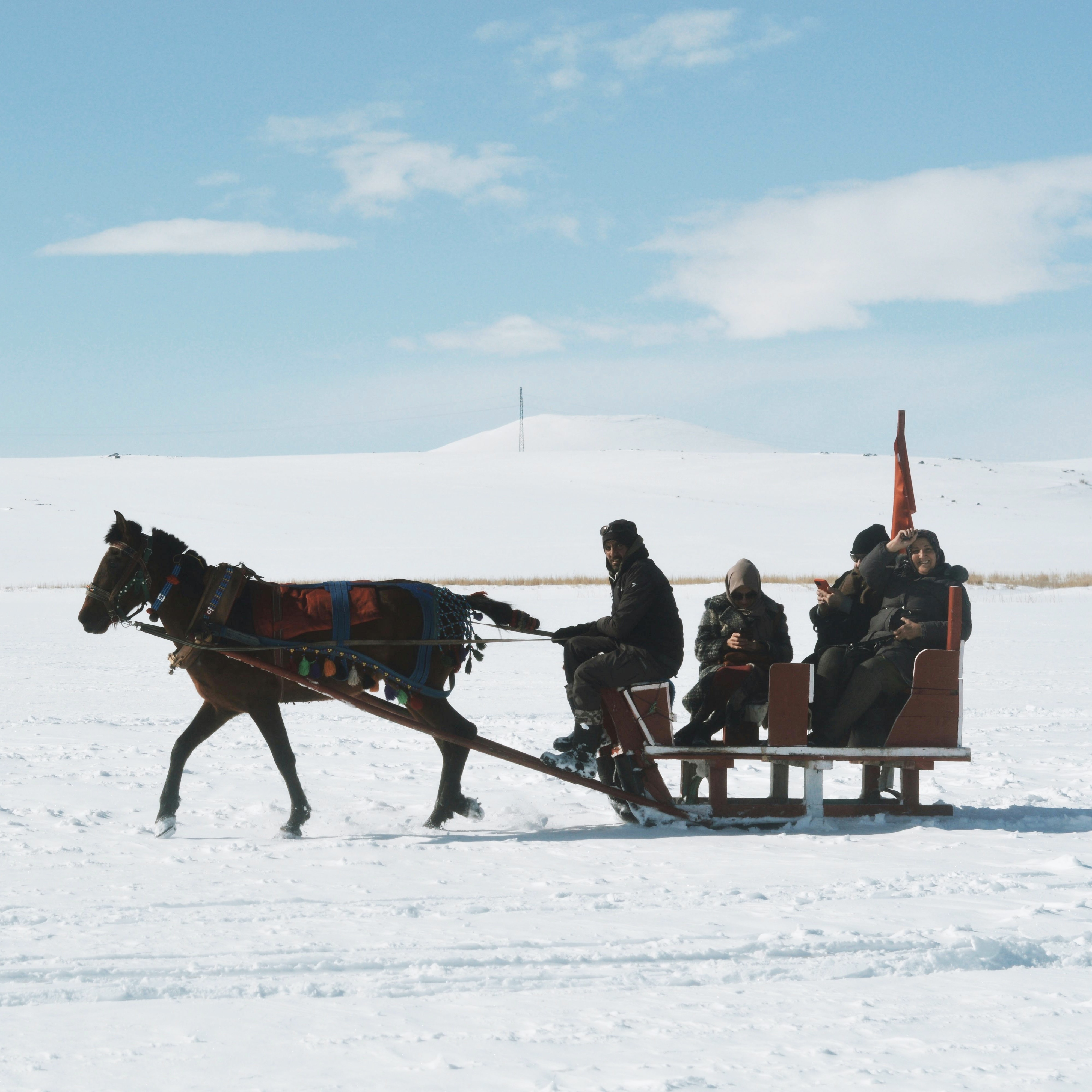}
    }\subfigure[Human Visual Saliency]{ 
        \includegraphics[width=0.4\columnwidth]{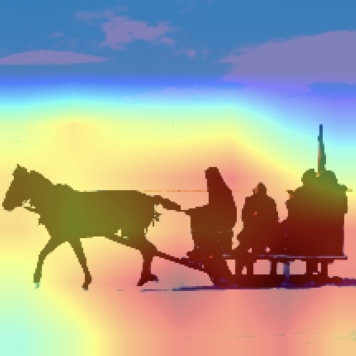}
    }
    \caption{\textbf{An example of human visual saliency analysis highlighting the areas of interest.} (a) illustrates an image of a horse-drawn sled carrying four individuals in a snowy landscape, and (b) presents the human visual saliency map generated by SalGAN \cite{pan2017salgan}, which highlights the regions of the image with high saliency and information density that attract human visual attention.}   
    \label{human-visual-saliency}
\end{figure}

To address the limitations in processing high-resolution images, several strategies have been proposed. Common approaches include training encoders to process larger images \cite{chen2022pali}, utilizing multiple encoders \cite{lu2024deepseek,cambrian}, and employing sub-image partitioning \cite{monkey,liu2024llavanext, chen2024farinternvl1_5}. However, training vision encoders for higher resolutions require significant computational resources and time with limited scalability, while using multiple encoders introduces architectural redundancy. As a result, sub-image partitioning has become the preferred approach. This method effectively mitigates visual information loss that occurs in fixed-resolution processing by dividing high-resolution images into smaller, manageable tiles that vision encoders can process. Sub-image partitioning provides a practical solution to overcome the constraints of fixed-resolution models, making it a more viable solution compared to alternatives.

Despite its advantages, sub-image partitioning has key limitations. 
When sub-images are processed, the varying information density within each sub-image poses another challenge. Conventional sub-image partitioning strategies typically fail to account for the differing information density across sub-images, applying uniform processing to all sub-images. This uniform treatment disregards the importance of key information, particularly in complex regions, thereby leading to the omission of critical details and producing suboptimal visual representations. This lack of adaptive processing limits the model’s ability to integrate local details and form a comprehensive understanding of high-resolution images.

\begin{figure}[h!]
\centering
\includegraphics[width=1\columnwidth]{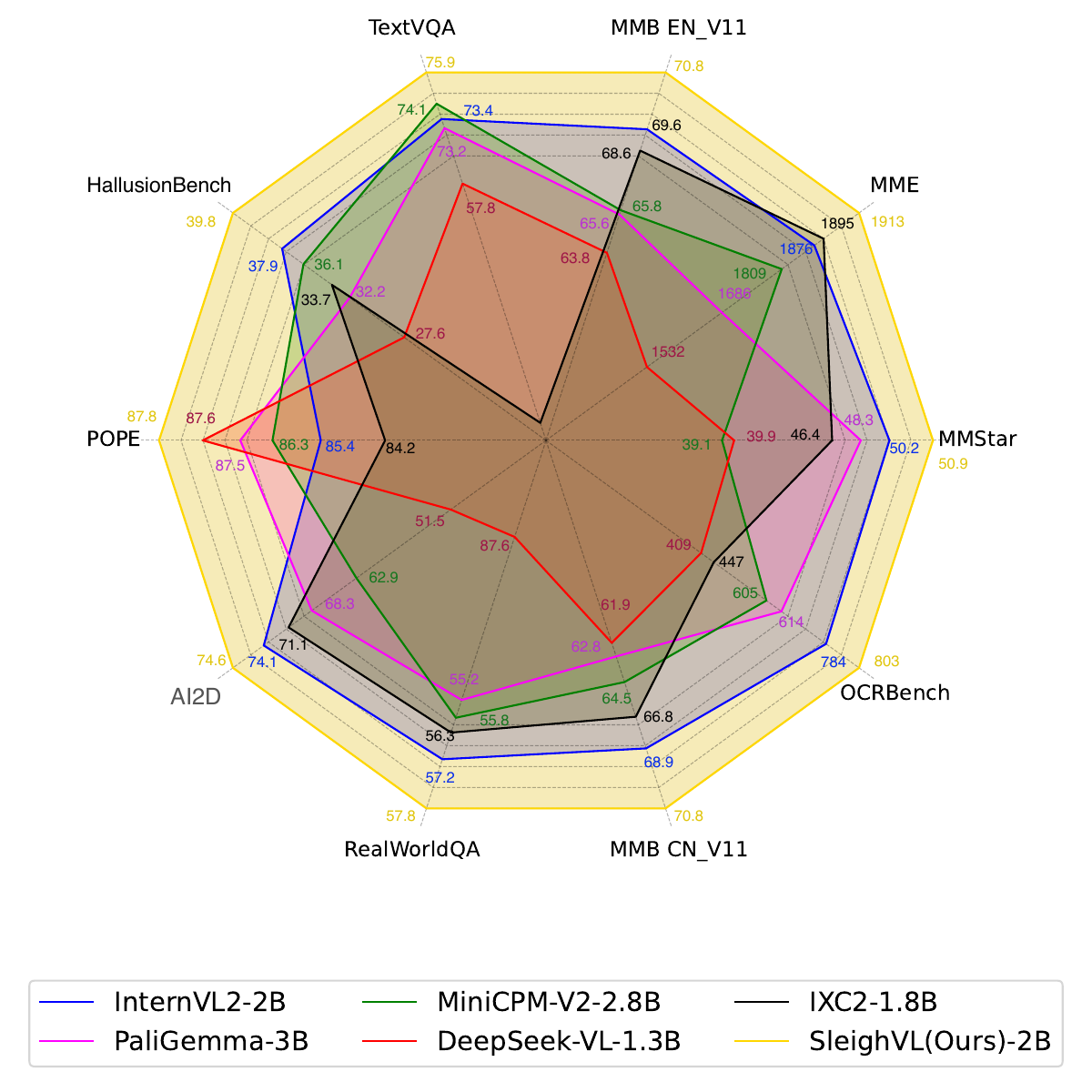}
\caption{\textbf{Radar chart comparing our model with existing popular LVLMs of similar parameter scales across ten benchmarks.}}
\label{Radar-Chart}
\end{figure}

Inspired by human visual saliency research \cite{pan2017salgan}, we recognize that when humans browse images, they tend to focus on regions with higher visual saliency in a bottom-up manner, as shown in Fig. \ref{human-visual-saliency}. 
When presented with an image, human eyes naturally gravitate toward information-dense or unusually prominent areas of an image. However, due to the inherent nature of sub-image partitioning, information-dense regions are inevitably fragmented across multiple sub-images and processed in a uniform manner. To address this limitation, we introduce the Global Semantic-guided Weight Allocator (GSWA) module, which is designed to mimic the bottom-up visual attention mechanism observed in human vision.  In the GSWA module, self-attention interactions between the most representative  $<cls>$  tokens of different sub-images, after ViT encoding, are employed to sense and quantify the relative information density of each sub-image. The GSWA module dynamically adjusts weights based on information densities across sub-images, improving local detail integration and overall image understanding. It enhances high-resolution image processing and boosting performance in complex scenarios. We also introduce a lightweight LVLM, SleighVL, by integrating our module to InternVL2-2B \cite{chen2024farinternvl1_5}  framework. SleighVL achieves significant performance across multiple benchmarks, as shown in Fig. \ref{Radar-Chart}.

{\color{black}In summary, our main contributions are as follows:} \begin{itemize}
    \item {\color{black} We conducted a comprehensive analysis of how sub-images from image partitioning carry varying levels of information densities and contribute differently to LVLM performance.}
    \item {\color{black} We proposed a lightweight module that dynamically assigns weights to sub-images based on their information density, allowing the model to focus on the most informative regions for better performance.}
    \item {\color{black} We integrated the proposed module into the InternVL2 framework and developed a lightweight yet high-performing LVLM, achieving significant improvements over the baseline model's performance. We also compare our method with existing state-of-the-art (SOTA) models across multiple benchmarks, consistently demonstrating outstanding results.}
\end{itemize}

\section{Related Work}
\subsection{Large Vision-Language Models}
Recent advancements in large language models (LLMs) have significantly improved natural language understanding and generation, facilitating breakthroughs in vision-language integration. This progress has led to the development of LVLMs, such as LLaVA \cite{liu2024llava}, InstructBLIP \cite{instructblip} and ShareGPT4V \cite{chen2025sharegpt4v}. These LVLMs typically consist of three key components: a vision encoder, a vision-language projector, and an LLM. The vision encoder, often based on Vision Transformer (ViT) \cite{clip,TMM_Vision_LanguagePre-Trained}, extracts visual features from the input images. The vision-language projector aligns the extracted visual features with the input embedding space of the LLM \cite{TMM_SgVA-CLIP}. The LLM then generates responses by jointly processing the aligned visual features and textual instructions \cite{TMM_GPT4Ego}.

However, initial LVLMs struggle with high-resolution images, as their resolution capacity is typically limited by the vision encoders. Vision encoders are optimized for lower resolutions, making it difficult to capture fine-grained details in high-resolution images. Our work enables the capabilities of LVLMs to process high-resolution images. By leveraging the fine-grained details inherent in such images, the model is able to achieve improved visual perception and a more comprehensive understanding of visual content.

\subsection{High-Resolution Processing in LVLMs}

To address the limitations of initial LVLMs in processing high-resolution images, several strategies have been proposed, including the training of vision encoders capable of handling higher resolutions, the use of multiple vision encoders operating in parallel, and sub-image partitioning techniques. Training higher-resolution vision encoders directly increases the resolution capacity of LVLMs, but training such encoders requires significant computational resources and time \cite{chen2022pali}. Additionally, the trained encoders are prone to visual information loss due to the constraints of fixed-resolution processing. Employing multiple encoders in parallel, with one branch processing high-resolution images and another handling low-resolution versions, also faces scalability issues while introducing architecture redundancy \cite{cambrian}. Sub-image partitioning techniques, on the other hand, divide the original image into smaller sub-images that can be processed by vision encoders, offering simplicity and high scalability \cite{chen2024farinternvl1_5, monkey}. However, when objects are split across multiple sub-images, spatial relationships may be lost, hindering the model’s ability to capture the full context. Although some works \cite{dong2024internlmxcomposer4khd,hu2024mplugDocOwl2} have sought to address this challenge, common sub-image partitioning algorithms still treat all sub-images equally, failing to focus on key areas of the image and distinguish regions containing critical information.

Our work enhances sub-image partitioning algorithms by introducing the GSWA module. By emulating how humans process images, the GSWA module dynamically assigns weights to sub-images during partitioning based on information density, helping the model focus on informative regions within the image. This approach improves the model’s visual understanding capabilities.

\section{Preliminary Observations}
\label{Section3}
One inherent characteristic of human visual cognition is the ability to selectively process visual signals, prioritizing regions with higher information density \cite{pan2017salgan}. This mechanism allows humans to focus on key areas that are critical for understanding the overall context. However, current sub-image partitioning-based LVLMs lack such a mechanism. Instead, they treat all sub-images equally, which limits their ability to achieve effective visual perception in complex scenarios. To address this limitation, we propose and verify two key hypotheses: 1) Sub-images that are more semantically aligned with the full image typically encapsulate richer visual information; 2) Sub-images with higher information density are more critical for supporting the model's visual understanding capabilities.

We adopt InternVL2-2B \cite{chen2024farinternvl1_5} as the baseline model to demonstrate the hypotheses. This model comprises an InternViT-300M \cite{chen2024internvl} (a CLIP-style vision encoder) for image encoding, an MLP for vision-language alignment, and an InternLM2-1.8B \cite{cai2024internlm2} for multimodal understanding and text generation. For any high-resolution image input, the model dynamically partitions the image into a set of sub-images, each with a resolution that meets the image encoder’s requirement (448px).

\begin{figure*}[t]
    \centering
    \subfigure[Football Scene]{
        \includegraphics[width=0.9\columnwidth]{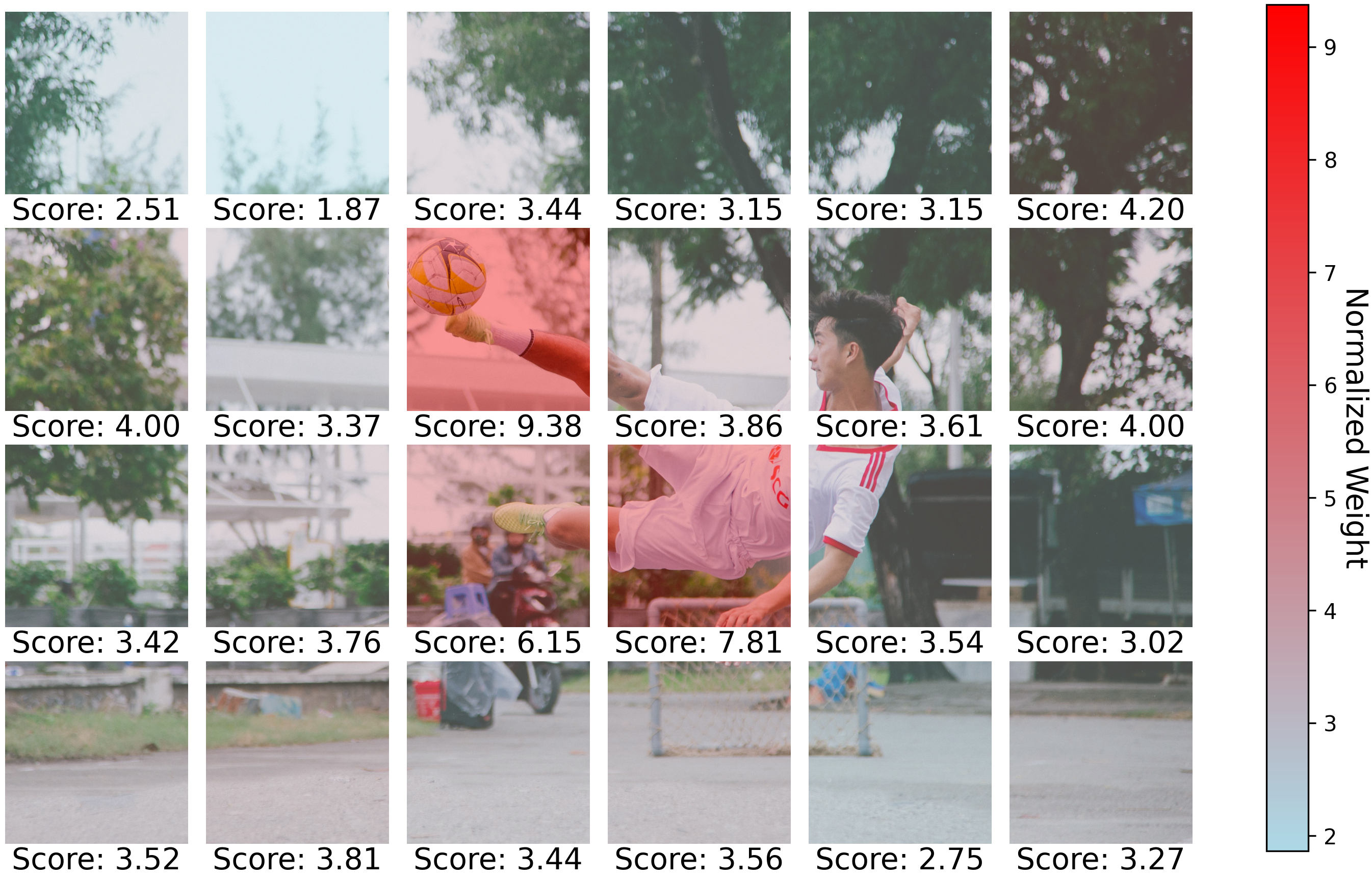}
    }\subfigure[Basketball Scene]{ 
        \includegraphics[width=0.9\columnwidth]{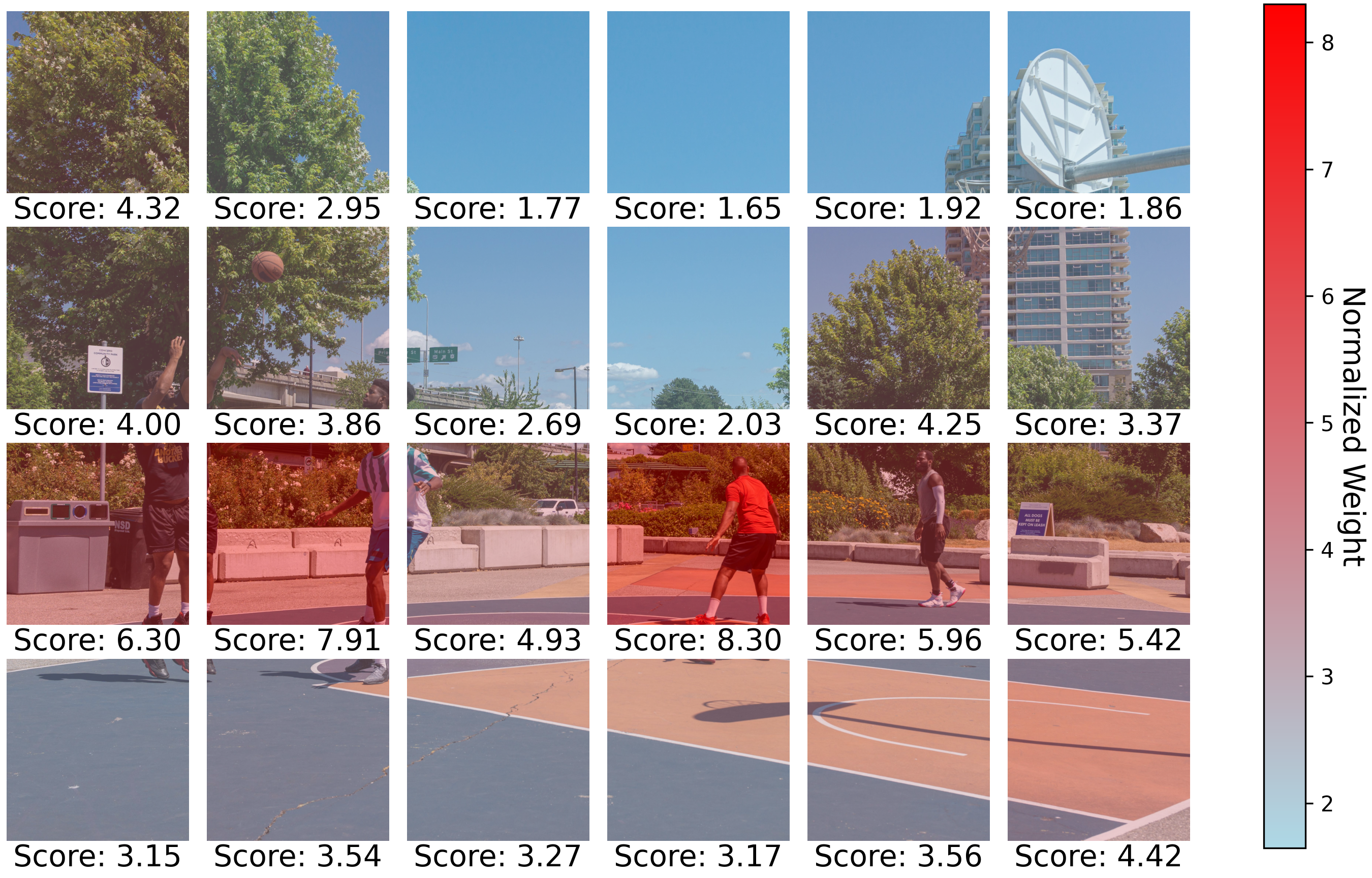}
    }
    \caption{\textbf{Two case studies to examine the semantic similarity distribution between each sub-image and the global image.} (a) illustrates the degree of similarity between each sub-image and the global image semantics "the player and football" in a football scene, and (b) illustrates the degree of similarity between each sub-image and the global image semantics "the basketball players and court" in a basketball scene.}   
    \label{case-study}
\end{figure*}
\begin{table*}[!t]
    \caption{\textbf{Preliminary observation results.} The baseline model (InternVL2-2B) is evaluated across six widely-adopted benchmarks under three distinct sub-image removal settings.}
    \label{model-comparison-observation}
    \centering
    \resizebox{1\textwidth}{!}{
    \begin{tabular}{l| c c c c c c c c}
        \hline
        \diagbox[innerwidth=3.5cm,innerleftsep=0.2cm,innerrightsep=1pt]{Method}{Benchmarks} & MME\cite{mme} & OCRBench\cite{ocrbench} & TextVQA\cite{textvqa} & CCBench\cite{mmbench_ccbench} & AI2D\cite{ai2d} & HallusionBench\cite{guan2023hallusionbench} \\ 
        \hline
        InternVL2-2B & 1876 & 784 & 73.4 & 74.8 & 74.1 & 37.9  \\
        Remove\_lowest\_3  & 1867 & 774 & 74.1 & 74.7 & 73.8 & 37.5 \\
        Remove\_top\_3  & 1734 & 440 & 69.1 & 69.4 & 71.2 & 35.6 \\
        Remove\_second-top\_3  & 1793 & 683 & 71.5 & 70.5 & 73.1 & 37.1 \\
         
         \hline
    \end{tabular}}
\end{table*}
To provide qualitative evidence for our first hypotheses, we conduct two case studies to examine the semantic similarity distribution between each sub-image and the global image. The sub-images are generated using the partitioning algorithm of the baseline model, and the semantic similarity scores are calculated as the cosine similarity between the $<cls>$ token of each sub-image and the $<cls>$ token of the resized full image. As shown in Fig. \ref{case-study}, sub-images with higher semantic relevance consistently correspond to regions containing key objects or contextual elements crucial for understanding the overall scene (e.g., the player and football in Fig. \ref{case-study} (a) or the basketball players and court in Fig. \ref{case-study} (b)). Conversely, sub-images with lower scores are primarily located at the periphery or dominated by background elements such as trees, sky, or distant buildings, contribute less to the semantic understanding of the scene. These findings qualitatively support our first hypothesis, demonstrating that sub-images more semantically aligned with the full image encapsulate richer visual information, making them essential for high-resolution image understanding.

To further reinforce the importance of global semantic relevance, we conduct a three-fold quantitative experiment. The baseline model is evaluated on six widely adopted benchmarks under three different sub-image removal settings: 1) removing the top 3 sub-images with the highest semantic similarity to the entire image; 2) removing the second top 3 sub-images with the highest semantic similarity to the entire image; and 3) removing the 3 sub-images with the lowest semantic similarity to the entire image. 

The results, summarized in Table \ref{model-comparison-observation}, reveal significant performance drops when sub-images with the highest global semantic similarity are removed. For instance, on the MME \cite{mme}  benchmark, the score decreases from 1876 to 1734, and on OCRbench \cite{ocrbench}, the accuracy significantly drops from 784 to 440. Similar trends are observed across other benchmarks. On the other hand, removing sub-images with the lowest semantic similarity results in minimal performance degradation. The MME score decreases only slightly from 1876 to 1867, and performance on OCRBench, CCBench \cite{mmbench_ccbench}, AI2D \cite{ai2d}, and HallusionBench \cite{guan2023hallusionbench} remains nearly unchanged. Notably, the score on TextVQA \cite{textvqa} even slightly improves, possibly because sub-images with the lowest global semantic relevance contain task-irrelevant noise. These results highlight a stark contrast in the impact of removing semantically important versus unimportant sub-images. Sub-images with higher semantic relevance to the entire image not only encapsulate richer visual information but also play a crucial role in preserving the model's ability to perform accurate visual perception and reasoning. These findings quantitatively validate our second hypothesis: sub-images with higher information are indeed more critical for supporting the model's visual understanding capabilities.

\begin{figure*}[ht!]
\centering
\includegraphics[width=0.7\textwidth]{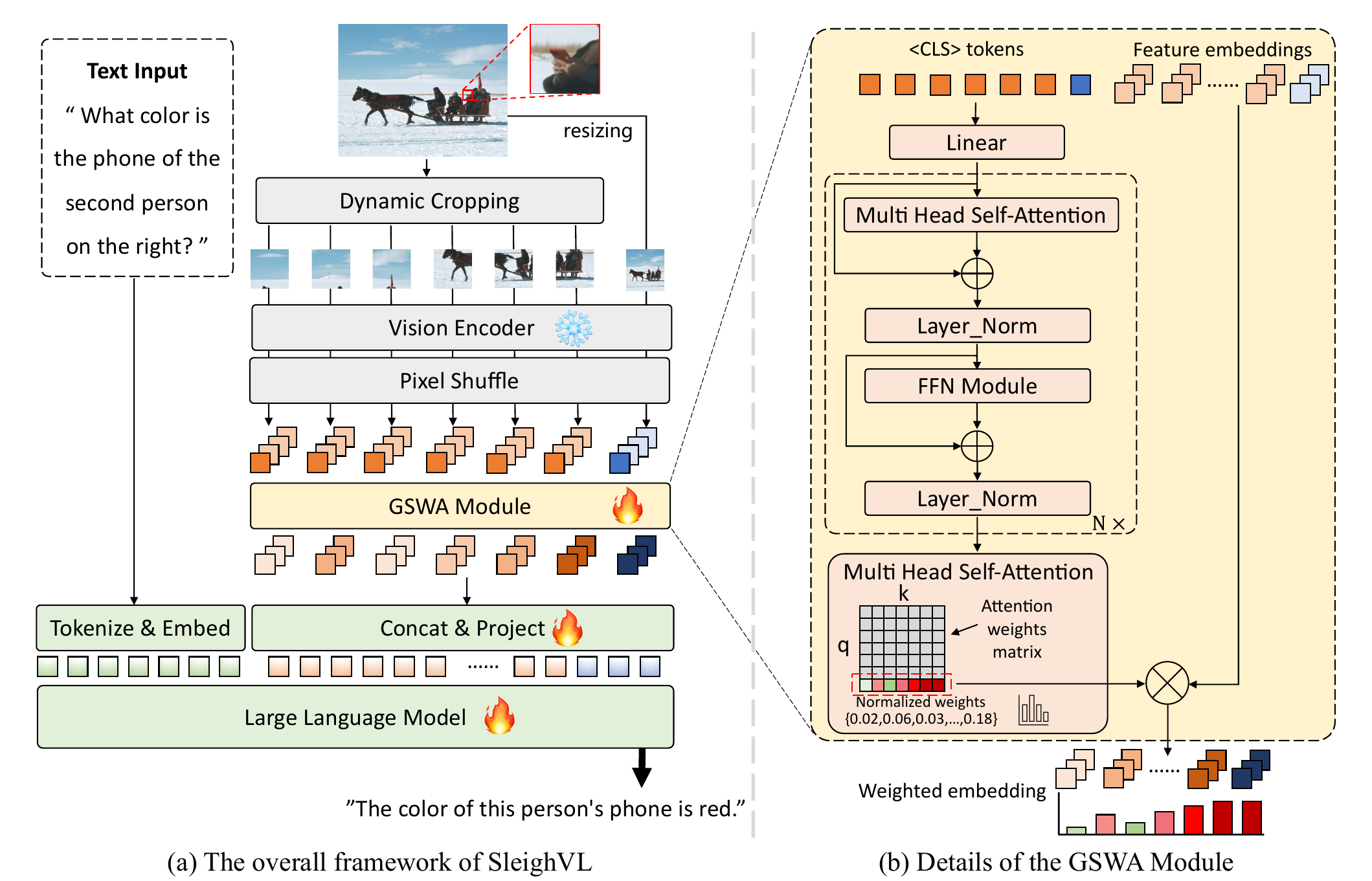}
\caption{\textbf{The workflow of the proposed method.} (a) describes the overall framework of SleighVL. (b) shows the design details of the global semantic-guided weight allocator.}
\label{Overall-frame}
\end{figure*}

\section{Methods}

Based on the two key insights outlined in Section \ref{Section3}, we posit that LVLMs should differentiate the treatment of sub-images based on their relative information density. To this end, we propose a Global Semantic-Guided  Weight Allocator (GSWA) module, which enables the model to autonomously perceive the information densities of different sub-images by quantitatively allocating corresponding weights, thereby addressing the limitation of uniform treatment of sub-images. As the proposed module can be seamlessly integrated into any LVLM that uses the sub-image partitioning strategy, we employ the widely-adopted InternVL2 \cite{chen2024farinternvl1_5} framework for demonstration in the following. 

\subsection{Overview}

As shown in Fig. \ref{Overall-frame}(a), the complete model architecture consists of five key components: a dynamic cropping module, a vision encoder, a GSWA module, a vision-language projector, and a large language model.  We provide a comprehensive overview of each module as follows.

\textbf{Dynamic Cropping module}. In our method, the preprocessing pipeline for high-resolution images follows the configuration established by InternVL2 \cite{chen2024farinternvl1_5}. To maintain the natural aspect ratio during processing, predefined aspect ratios are determined based on the specified maximum number of sub-images. During the matching process, the aspect ratio of the input image is calculated and compared to the absolute differences with the predefined ratios. 



\textbf{Vision Encoder.} We employ Intern-ViT \cite{chen2024internvl} as the vision encoder, which accepts a fixed input image resolution of $448\times448$. The input image $I \in \mathbb{R}^{H \times W \times C}$, is passed through the vision encoder for feature extraction. After the input image is cropped into $N$ sub-images along with one thumbnail representing the global image, they are then passed through multiple transformer layers to ultimately obtain the corresponding visual embeddings:
\begin{equation}
\textbf{F} = VisionEncoder(Crop(\textbf{I})) \in \mathbb{R} ^ {(N+1) \times (M+1) \times D},
\end{equation}
where $M+1$ and $D$ denote the number and dimension of the extracted visual embeddings for each sub-image, respectively. It is important to note that each set of visual embeddings comprises M patch features accompanied by a $<cls>$ token, which contains the global semantic information of the image. Meanwhile, the output from the vision encoder will be processed through a pixel shuffle mechanism:
\begin{equation}
\textbf{F}^{'} = PixelShuffle(\textbf{F}) \in \mathbb{R} ^ {(N+1) \times (M/4+1) \times 4D},
\end{equation}
reducing the number of visual embeddings to one quarter and quadrupling the dimension of the extracted visual embeddings.

\textbf{Global Semantic-guided Weight Allocator.} This module is designed to leverage the semantic information of the global image to guide the weight allocation of sub-images, enabling the adaptive assignment of different weight values to sub-images based on information densities. The module receives $N+1$ sets of visual embeddings. After taking interactions of the $<cls>$ token in each set, the GSWA module perceives the information density of the sub-images and weights the feature embedding sets:
\begin{equation}
\textbf{F}^{'}_{weighted} = GSWA(\textbf{F}^{'}) \in \mathbb{R} ^ {(N+1) \times (M/4) \times 4D}.
\end{equation}
Following the design of InternVL2 \cite{chen2024farinternvl1_5}, $<cls>$ tokens are not passed to the subsequent stage. The details are provided in Section \ref{section42}. 

\textbf{Vision-Language Projector.} After feature extraction and weight allocation, the resulting visual embeddings are aligned with the input embedding space of LLM through the vision-language projector. Following the mainstream design \cite{llava1.5,liu2024llavanext,chen2024internvl}, the projector consists of two linear layers and a GELU activation function, providing an efficient and straightforward alignment:
\begin{equation}
\begin{aligned}
 Proj(\textbf{F}^{'}_{weighted}) = MLP(\hat{\textbf{F}}) \in \mathbb{R} ^ {(N+1) \times (M/4) \times D_t},
 \end{aligned}
\end{equation}
where $D_t$ is the hidden dimension of the LLM.

\textbf{Large Language Model.} The large language model first tokenizes and embeds the textual instructions. The textual embeddings are concatenated along the sequence dimension with the visual embeddings processed by the vision-language projector to form a unified input feature sequence. This combined sequence is then passed through multiple transformer layers, enabling the model to generate the response corresponding to the given multi-modal context.

\subsection{Global Semantic-guided Weight Allocator}
\label{section42}
As shown in Fig. \ref{Overall-frame}(b), the GSWA module comprises a linear layer, a stack of transformer blocks, and an additional multi-head self-attention layer. Each transformer block consists of a multi-head self-attention module and a feed-forward neural network, combined with two residual connections and two layer normalization modules.

Specifically, the GSWA receives $N+1$ visual embedding sets output from the pixel shuffle module. It then extracts the $<cls>$ tokens which capture the most representative features of the N sub-images and the global image. These tokens are mapped through a linear layer to the hidden dimension of the transformer blocks:

\begin{equation}
\textbf{F}^{'}_{cls} = Linear(\textbf{cls}_{1}, ...,\textbf{cls}_{{N}}, \textbf{cls}_{{global}}) \in \mathbb{R}^{(N+1) \times D_g},
\end{equation}

where $D_g$ is the hidden dimension of the GSWA. The mapped $<cls>$ tokens are then processed through stacked transformer blocks to obtain $N+1$ outputs that have perceived the varying degrees of information densities among the different sub-images:

\begin{equation}
\textbf{F}^{''}_{cls} = \mathit{Transformers}(\textbf{F}^{'}_{cls}) \in \mathbb{R}^{(N+1) \times D_g}.
\end{equation}

Subsequently, these outputs are fed into a multi-head self-attention layer to obtain the average attention weights across k attention heads. Each row of the attention weight matrix is a normalized result, using the global $<cls>$ token’s attention assigned to the sub-image $<cls>$ tokens and itself as the final weights to be allocated:

\begin{equation}
\begin{split}
   \mathbf{w} = \{\frac{1}{k} \sum_{h=1}^{k} \text{softmax}\left( \frac{\mathbf{F}''_{\text{cls}} \mathbf{W}_h^Q \left( \mathbf{F}''_{\text{cls}} \mathbf{W}_h^K \right)^\top}{\sqrt{D_g}} \right )\}_{global}, 
\end{split}
\end{equation}
where $\mathbf{w} \in \mathbb{R}^{N+1}$, $ \sum_{n=1}^{N+1} \mathbf{w_n} = 1$, $k$ is the number of attention heads in the multi-head self-attention mechanism, $\mathbf{W}_h^Q$ and $\mathbf{W}_h^K$ is the query and key projection matrix for the $h$-th attention head respectively, $\{\cdot\}_\text{global}$ is an operation that extracts the average attention weights assigned by the global $<cls>$ token to the $<cls>$ token of each sub-image.

Finally, the $N+1$ sets of visual embeddings are weighted accordingly, serving as the final output of the GSWA module.
\begin{equation}
\begin{aligned}
\textbf{F}^{'}_{weighted} = \{\mathbf{w_1} \times \textbf{F}^{'(1)}_{patch}, \mathbf{w_2} \times \textbf{F}^{'(2)}_{patch}, ..., \\ \mathbf{w_N} \times \textbf{F}^{'(N)}_{patch},  \mathbf{w_{global}} \times \textbf{F}^{'(global)}_{patch}\},
\end{aligned}
\end{equation}
where $\textbf{F}^{'(n)}_{patch}$ represents the visual embeddings of the $n$-th sub-image with the $<cls>$ token removed, $\mathbf{w_n}$ represents the weight assigned by the GSWA module to the $n$-th sub-image, $\mathbf{w_{global}}$ represents the weight of the global image thumbnail, and $\textbf{F}^{'(global)}_{patch}$ represents the visual embeddings of the global image thumbnail after the removal of the $<cls>$ token.

Through the GSWA module, each sub-image's visual embeddings are weighted. After fine-tuning, the model can quantitatively perceive the information density of each sub-image, thereby better focusing on the sub-images containing critical features.

\section{Experiments}
\subsection{Implementation Details}
\textbf{Model Architecture}. We employ InternVL2-2B \cite{chen2024farinternvl1_5} as the baseline framework to demonstrate the effectiveness of our method.  The final model includes an InternLM2-1.8B \cite{cai2024internlm2} LLM backbone, a two-layer MLP projector, an InternViT-300M \cite{chen2024internvl} vision encoder, and a GSWA module. The GSWA module consists of four stacked transformer blocks with a hidden dimension of 1024, and each multi-head self-attention layer contains four attention heads. The additional multi-head self-attention layer follows the same configuration. 

\textbf{Training Hyperparameter}. We set the maximum number of sub-images for the dynamic cropping algorithm to 8 and the drop path rate of the vision encoder to 0.1. AdamW \cite{adamw} was used as the optimizer with a learning rate of 4e-9, following a cosine decay schedule with a warmup ratio of 0.03 and a weight decay of 0.01. 

\textbf{Training Data}. The fine-tuning training data for the model includes AI2D \cite{ai2d}, Llava-150K-zh \cite{llava1.5}, DVQA \cite{kafle2018dvqa}, ChartQA \cite{chartqa}, DocVQA \cite{docvqa}, GeoQA+ \cite{GeoQa+}, Synthdog-en \cite{Synthdogen}, and Sharegpt4V-100K \cite{chen2025sharegpt4v}, comprising a total of 602K training samples.

\textbf{Benchmarks}. We compare our SleighVL with SOTA open-source LVLMs by categorizing them into two groups based on model size: models with $<4$ billion parameters and models with $\geq4$ billion parameters. The comparison spans four categories, encompassing a total of 21 benchmarks. The General category includes MME \cite{mme}, MMbench \cite{mmbench_ccbench} (MMBench-EN, MMBench-CN, MMBench-EN-V11, MMBench-CN-V11), MMStar\cite{chen2024wemmstar}, MM-Vet \cite{mmvet}, and SEED-Image \cite{seed}; The Realworld category comprises RealWorldQA \cite{realworldqa}, MME-RealWorld \cite{zhang2024mmerealworld}, HRBench4K \cite{hrbench}, and HRBench8K \cite{hrbench}; The Text-rich VQA category consists of InfoVQA \cite{mathew2022infographicvqa}, TextVQA \cite{textvqa}, OCRBench \cite{ocrbench}, and DocVQA \cite{docvqa}; The Others category includes HallusionBench \cite{guan2023hallusionbench} and POPE \cite{pope} for hallucination testing, ScienceQA \cite{sqa} and AI2D \cite{ai2d} for scientific question answering, and CCBench \cite{mmbench_ccbench} for evaluating Chinese scenario capabilities. 


\subsection{General Benchmark Results}
\begin{table*}[t]
\centering
\caption{\textbf{Comparison to different LVLMs on General Benchmarks.} The models are categorized into two groups: parameter scale less than 4 billion and greater than or equal to 4 billion. The \textbf{bold} values represent the best scores in each scale group. The MME scores here are the sum of perception and cognition scores. }
\label{model-performance-general}
\renewcommand{\arraystretch}{} 
\resizebox{0.8\textwidth}{!}{%
\begin{tabular}{l | c | c c c c c c c c}
\toprule
Models & \#Param & MME & \makecell[c]{MMB\\EN} & \makecell[c]{MMB\\EN\_V11} & \makecell[c]{MMB\\CN} & \makecell[c]{MMB\\CN\_V11} & \makecell[c]{MM\\Star} & MM-Vet & SEED-I \\
\midrule
Monkey \cite{monkey} & 9.8B & 1888 & 72.4 & 67.3 & 67.5 & 65.1 & 40.7 & 41.0 & 68.9 \\
Cambrian-8B \cite{cambrian} & 8B & 1803 & 75.9 & 68.2 & 67.9 & 64.9 & 50.7 & \textbf{48.0} & \textbf{74.7} \\
IDEFICS2-8B \cite{2024idefics2} & 8B & 1848 & 76.8 & 68.9 & 68.6 & 65.1 & 49.5 & 34.0 & 71.9 \\
DeepSeek-VL-7B \cite{lu2024deepseek} & 7.3B & 1766 & 73.2 & 70.7 & 71.4 & 69.7 & 40.5 & 41.5 & 70.4 \\
VILA1.5-13B \cite{lin2023vila} & 13B & 1719 & 74.4 & 68.5 & 67.7 & 65.2 & 44.2 & 45.0 & 72.6 \\
ShareGPT4V-7B \cite{chen2025sharegpt4v} & 7.2B & 1944 & 68.8 & 61.6 & 62.2 & 58.6 & 35.7 & 37.6 & 69.7 \\
InstructBLIP-7B \cite{instructblip} & 8.2B & 1392 & 33.9 & 23.9 & 38.4 & 58 & 32.7 & 33.1 & 44.5 \\
LLaVA-V1.5-7B \cite{llava1.5} & 7.2B & 1809 & 64.3 & 53.1 & 58.3 & 43.9 & 31.1 & 32.9 & 66.1 \\
MiniCPM-V2.5 \cite{yao2024minicpm} & 8B & \textbf{2025} & \textbf{77.6} & \textbf{72.0} & \textbf{73.8} & \textbf{70.1} & \textbf{51.8} & 52.8 & 72.3 \\
\hline
InternVL2-2B \cite{chen2024farinternvl1_5}& 2B & 1876 & 73.2 & 69.6 & 70.9 & 68.9 & 50.2 & 39.6 & 70.9 \\
VILA1.5-3B \cite{lin2023vila} & 3B & 1647 & 64.5 & 58.8 & 56.2 & 54.5 & 40.6 & 38.8 & 68.0 \\
MiniCPM-V2 \cite{yao2024minicpm} & 2.8B & 1809 & 69.1 & 65.8 & 66.5 & 64.5 & 39.1 & 41.0 & 67.1 \\
IXC2-1.8B \cite{dong2024internlm_xc2} & 2B & 1895 & 73.0 & 68.6 & 68.4 & 66.8 & 46.4 & 31.2 & 70.8 \\
DeepSeek-VL-1.3B \cite{lu2024deepseek} & 2B & 1532 & 64.6 & 63.8 & 62.9 & 61.9 & 39.9 & 34.8 & 66.7 \\
PaliGemma-448px \cite{beyer2024paligemma} & 3B & 1686 & 71.0 & 65.6 & 63.6 & 62.8 & 48.3 & 33.1 & 69.6 \\
LLaVA-OV-0.5B \cite{li2024llavaonevision} & 1B & 1478 & 61.6 & 56.8 & 55.5 & 53.9 & 37.5 & 29.1 & 65.5 \\
\textbf{SleighVL(Ours)} & 2B & \textbf{1913} & \textbf{73.4} & \textbf{70.8} & \textbf{77.2} & \textbf{70.8} & \textbf{50.9} & \textbf{42.1} & \textbf{71.2} \\
\bottomrule
\end{tabular}
}
\end{table*}

The experiment results clearly show that our proposed model exhibits outstanding performance and demonstrate its comprehensive strength across general tasks, as summarized in Table \ref{model-performance-general}. SleighVL not only outperforms models with fewer than 4 billion parameters but also demonstrates strong competitiveness when compared to models with more than 4 billion parameters.

In the category of models with fewer than 4 billion parameters, SleighVL scores 1913 in MME, surpassing the baseline model InternVL2-2B (1876), as well as other competing models like MiniCPM-V2 \cite{yao2024minicpm} (1809) and InternLM-XComposer2-1.8B \cite{dong2024internlm_xc2} (1895).  Additionally, SleighVL also exhibits broad advantages across other comprehensive evaluation benchmarks, such as MMB\_CN and SEEDBench-Image, with scores of 77.2 and 71.2, respectively, further highlighting its adaptability and consistency in diverse evaluation scenarios.

Notably, SleighVL still exhibits highly competitive performance when compared to models with larger parameter scales. On MMB\_EN and MMB\_CN\_V11, SleighVL (73.4/70.8) not only closely approaches but even surpasses several larger models, such as Monkey \cite{monkey} (72.4/65.1), Cambrian-8B \cite{cambrian} (75.9/64.9), and IDEFICS2-8B \cite{2024idefics2} (76.8/65.1). In challenging benchmarks such as MMStar and MM-Vet, SleighVL maintains its lead with scores of 50.9 and 42.1, significantly outperforming many models with larger parameters, including Monkey, IDEFICS2-8B, and DeepSeek-VL-7B \cite{lu2024deepseek}.  

\subsection{Realworld Benchmark Results}

\begin{table*}[t]
\centering
\caption{\textbf{Comparison to different LVLMs on Realworld Benchmarks.} The P and R in MME-RealWorld means perception and reasoning respectively.}
\label{model-performance-realworldqa}
\renewcommand{\arraystretch}{} 
\resizebox{0.8\textwidth}{!}{%
\begin{tabular}{l | c | c  c  c  c}
\toprule
Models & \#Param & RealWorldQA & MME-RealWorld(P/R) & HRbench4K & HRbench8K \\
\midrule
IXC2-4KHD \cite{dong2024internlmxcomposer4khd} & 7B & 63.3 & - & \textbf{57.8} & \textbf{51.3}  \\
IXC2-7B \cite{dong2024internlm_xc2} & 7B & \textbf{63.8} & - & 46.0 & 37.9 \\
MiniCPM-V2.5 \cite{yao2024minicpm} & 8B & 63.5 & \textbf{44.0} / 35.9 & - & - \\
Cambrian-8B \cite{cambrian} & 8B & 64.2 & 43.5 / \textbf{38.4} & - & - \\
DeepSeek-VL-7B \cite{lu2024deepseek} & 7.3B & 54.2 & 31.7 / 28.2 & 35.5 & 33.4 \\
LLaVA-v1.6-7B \cite{liu2024llavanext} & 7.1B & 57.8 & - & 47.9 & 40.8  \\
Monkey \cite{monkey} & 9.8B & 52.4 & 32.8 / 27.0 & - & - \\
LLaVA-V1.5-13B \cite{llava1.5} & 13.4B & 55.3 & 26.8 / 25.8 & 43.3 & 37.8 \\
\hline
InternVL2-2B \cite{chen2024farinternvl1_5} & 2B & 57.2 & 40.1 / 33.0 & 52.7 & 42.5 \\
DeepSeek-VL-1.3B \cite{lu2024deepseek} & 2B & 49.7 & - & - & - \\
Mini-InternVL-V1.5 \cite{gao2024miniinternvl} & 2B & \textbf{57.9} & 39.9 / 30.1 & - & - \\
MiniCPM-V2 \cite{yao2024minicpm} & 2.8B & 55.8 & - & - & - \\
PaliGemma-448px \cite{beyer2024paligemma} & 3B & 55.2 & - & - & - \\
IXC2-1.8B \cite{dong2024internlm_xc2} & 2B & 56.3 & - & - & - \\
LLaVA-OV-0.5B \cite{li2024llavaonevision} & 1B & 55.6 & - & - & - \\
\textbf{SleighVL(Ours)} & 2B & 57.8 & \textbf{43.4} / \textbf{35.1} & \textbf{54.0} & \textbf{49.0} \\
\bottomrule
\end{tabular}
}
\end{table*}

Table \ref{model-performance-realworldqa} corresponds to the performance comparison under real-world scenarios. Among the benchmarks, HRbench4K and HRbench8K are specially designed to evaluate the visual understanding capabilities for high-resolution images.

In the category of models with fewer than 4 billion parameters, SleighVL leads comprehensively in overall performance. In RealWorldQA, SleighVL achieves a score of 57.8, surpassing the baseline model InternVL2-2B (57.2), as well as competing models such as MiniCPM-V2 (55.8) and PaliGemma-448px \cite{beyer2024paligemma} (55.2). Additionally, in MME-RealWorld, SleighVL achieves 43.4/35.1 in Perception/Reasoning, significantly outperforming InternVL2-2B’s 40.1/33.0 and Mini-InternVL-V1.5’s 39.9/30.1 \cite{gao2024miniinternvl}.

In the category of models with more than 4 billion parameters, SleighVL achieved a score of 57.8 in RealWorldQA, surpassing models with larger parameters such as DeepSeek-VL-7B (54.2) and LLaVA-V1.5-13B \cite{llava1.5} (55.3), and approaching InternLM-XComposer2-4KHD \cite{dong2024internlmxcomposer4khd} (63.3). In HRbench4K and HRbench8K, SleighVL scored 54 and 49, significantly outperforming models such as InternLM-XComposer2-7B (46/37.9), DeepSeek-VL-7B (35.5/33.4). These results clearly demonstrate the effectiveness of the proposed GSWA module.

\subsection{Text-rich VQA Benchmark Results}

\begin{table*}[!t]
\centering
\caption{\textbf{Comparison to different LVLMs on Text-rich VQA Benchmarks.} Test and Val means the data split used in evaluation.}
\label{model-performance-VQA}
\renewcommand{\arraystretch}{} 
\resizebox{0.75\textwidth}{!}{%
\begin{tabular}{l | c | c  c  c  c}
\toprule
Models & \#Param & $\text{InfoVQA}^{Test}$ & $\text{TextVQA}^{Val}$ & OCRBench & $\text{DocVQA}^{Test}$ \\
\midrule
InternVL2-4B \cite{chen2024farinternvl1_5} & 4B & 67 & 74.7 & 788 & 89.2 \\
Monkey \cite{monkey} & 9.8B & 36.1 & 67.6 & 534 & 66.5 \\
IDEFICS2-8B \cite{2024idefics2} & 8B & - & 73.0 & 626 & 74.0  \\
Ovis1.6-9B \cite{lu2024ovis} & 10.2B & - & \textbf{78.2} & \textbf{830} & -  \\
VILA1.5-13B \cite{lin2023vila} & 13B & 30.4 & 65.0 & 460 & 58.6 \\
MiniCPM-V2.5 \cite{yao2024minicpm} & 8B & - & 76.6 & 725 & 84.8  \\
IXC2-4KHD \cite{dong2024internlmxcomposer4khd} & 7B & 68.6 & 77.2 & 675 & \textbf{90.0} \\
Cambrian-8B \cite{cambrian} & 8B & - & 71.7 & 624 & 77.8 \\
LLaVA-V1.5-7B \cite{llava1.5} & 7.2B & - & 58.2 & 318 & -  \\
LLaVA-OV-7B \cite{li2024llavaonevision} & 8B & \textbf{68.8} & - & 622 & 87.5 \\
\hline
InternVL2-2B \cite{chen2024farinternvl1_5} & 2B & 58.9 & 73.4 & 784 & 86.9 \\
MiniCPM-V2 \cite{yao2024minicpm} & 2.8B & - & 74.1 & 605 & 71.9 \\
Mini-InternVL-V1.5 \cite{gao2024miniinternvl} & 2B & 55.4 & 70.5 & 654 & 85.0 \\
DeepSeek-VL-1.3B \cite{lu2024deepseek} & 2B & - & 57.8 & 409 & - \\
PaliGemma-448px \cite{beyer2024paligemma} & 3B & 40.5 & 73.2 & 614 & 78.0 \\
LLaVA-OV-0.5B \cite{li2024llavaonevision} & 1B & 41.8 & 66.3 & 583 & 70.0 \\
\textbf{SleighVL(Ours)} & 2B & \textbf{59.3} & \textbf{75.9} & \textbf{803} & \textbf{87.1} \\
\bottomrule
\end{tabular}
}
\end{table*}

LVLMs are frequently employed in text-rich Visual Question Answering (VQA) tasks, which impose higher demands on the fine-grained visual understanding capabilities. Table \ref{model-performance-VQA} demonstrates the model comparison across four text-rich VQA benchmarks. 

SleighVL achieved comprehensive superiority among models with parameter scales below 4 billion. Notably, on OCRBench, it attained an outstanding score of 803, significantly surpassing the baseline model InternVL2-2B’s score of 784, and outperforming PaliGemma-448px (614) by 189 points. In TextVQA and DocVQA, SleighVL achieves scores of 75.9 and 87.1, markedly surpassing MiniCPM-V2’s scores of 74.1 and 71.9.

 SleighVL also comprehensively outperforms the models with larger scales, narrowing the gap with the top performers. On OCRBench, SleighVL achieves an outstanding score of 803, surpassing all models except Ovis1.6-9B \cite{lu2024ovis} (830). On TextVQA, SleighVL scores 75.9, significantly exceeding InternVL2-4B’s score of 74.7. For DocVQA, SleighVL achieved scores of 87.1, approaching those of LLaVA-OneVision-7B \cite{li2024llavaonevision} (87.5).

\subsection{Others Benchmark Results}

\begin{table*}[!t]
\centering

\caption{\textbf{Comparison to different LVLMs on Others benchmarks, including hallucination testing, scientific VQA, and Chinese VQA.}}
\label{model-performance-others}
\renewcommand{\arraystretch}{} 
\resizebox{0.7\textwidth}{!}{%
\begin{tabular}{l | c | c c | c c | c}
\toprule
Models & \#Param & HallusionBench & POPE & AI2D & SQA-I & CCBench \\
\midrule
Monkey \cite{monkey} & 9.8B & 39.3 & 83.5 & 62.6 & 69.4 & 48.0 \\
MiniCPM-V2.5 \cite{yao2024minicpm} & 8B & 42.4 & 86.7 & 78.4 & 89.2 & 45.9 \\
IXC2-4KHD \cite{dong2024internlmxcomposer4khd} & 7B & \textbf{42.5} & - & \textbf{81.0} & \textbf{96.3} & 47.1 \\
IDEFICS2-8B \cite{2024idefics2} & 8B & 39.1 & 86.2 & 72.3 & 88.7 & 37.6 \\
Cambrian-13B \cite{cambrian} & 13B & 39.4 & 86.8 & 73.6 & 79.3 & 26.7 \\
Cambrian-8B \cite{cambrian} & 8B & 30.6 & 86.4 & 73.0 & 80.4 & 23.7 \\
DeepSeek-VL-7B \cite{lu2024deepseek} & 7.3B & 34.5 & \textbf{88.1} & 65.3 & 80.9 & \textbf{52.4} \\
VILA1.5-13B \cite{lin2023vila} & 13B & 39.3 & 85.0 & 79.9 & 79.1 & 24.3 \\
ShareGPT4V-7B \cite{chen2025sharegpt4v} & 7.2B & 28.6 & 86.6 & 58.0 & 68.4 & 30.8 \\
InstructBLIP-7B \cite{instructblip} & 8.2B & 31.2 & 86.1 & 40.6 & 54.1 & 12.7 \\
LLaVA-V1.5-7B \cite{llava1.5} & 7.2B & 27.6 & 85.9 & 55.5 & 66.8 & 17.8 \\
\hline
InternVL2-2B \cite{chen2024farinternvl1_5} & 2B & 37.9 & 85.4 & 74.1 & 94.1 & 74.7 \\
VILA1.5-3B \cite{lin2023vila} & 3B & 31.2 & 86.8 & 57.9 & 69.2 & 24.1 \\
MiniCPM-V2 \cite{yao2024minicpm} & 2.8B & 36.1 & 86.3 & 62.9 & 80.7 & 45.3 \\
IXC2-1.8B \cite{dong2024internlm_xc2} & 2B & 33.7 & 84.2 & 71.1 & 92.2 & 40.4 \\
DeepSeek-VL-1.3B \cite{lu2024deepseek} & 2B & 27.6 & 87.6 & 51.5 & 68.4 & 37.6 \\
PaliGemma-448px \cite{beyer2024paligemma} & 3B & 32.2 & 87.5 & 68.3 & 94.3 & 29.6 \\
LLaVA-OV-0.5B \cite{li2024llavaonevision} & 1B & 27.9 & \textbf{87.8} & 57.1 & 67.2 & 28 \\
\textbf{SleighVL(Ours)} & 2B & \textbf{39.8} & \textbf{87.8} & \textbf{74.6} & \textbf{94.4} & \textbf{75.7} \\
\bottomrule
\end{tabular}
}
\end{table*}

We further explore the model’s capabilities across several other dimensions, as shown in Table \ref{model-performance-others}. Specifically, we employe HallusionBench and POPE to assess hallucination detection ability, AI2D and SQA-Image to gauge scientific knowledge comprehension, and CCBench to evaluate Chinese question-answering capability.

Among the models with fewer than 4 billion parameters, SleighVL consistently maintains leading performance across all the benchmarks. On HallusionBench and POPE, our model achieved scores of 39.8 and 87.8 respectively, surpassing the baseline model InternVL2-2B’s scores of 37.9 and 86.8. In scientific VQA tasks, SleighVL scored 94.4 on SQA-Image, exceeding the second-place PaliGemma-448px’s score of 94.3 and the baseline model’s score of 94.1. For Chinese VQA tasks, our model boosts the baseline model's score from 74.7 to 75.7, significantly outperforming other models.

When comparing with larger models, SleighVL outperforms most of them and approaches the top-performing model. On HallusionBench, POPE, and AI2D, SleighVL (39.8/87.8/74.6) comprehensively exceeded the competing models, such as Cambrian-13B (39.4/86.8/73.6) and IDEFICS2-8B (39.1/86.2/88.7).  Additionally, our method nearly matched IXC2-4KHD’s score of 96.3 on SQA-Image.

\subsection{Ablation Study}

In this section, we demonstrate the effectiveness of our GSWA module through ablation experiments. As shown in Table \ref{model-comparison-ablation}, the ablation experiments progressively alter the settings of the core GSWA module within our model architecture, ultimately eliminating the module entirely from the model. 


\begin{table*}[!t]
    \caption{\textbf{Ablation study of weight allocation scheme in GSWA designing.} All configuration changes involve fine-tuning the baseline model. $\dag$ denotes the direct fine-tuning of the model.}
    \label{model-comparison-ablation}
    \centering
    \renewcommand{\arraystretch}{1.2} 
    \resizebox{1\textwidth}{!}{
    \begin{tabular}{l| c c | c c | c c | c c | c}
        \hline
        \diagbox[]{Method}{Benchmarks} & MME & MMStar & RealWorldQA & HRBench4K & OCRBench & $\text{TextVQA}^{Val}$  & HallusionBench & CCBench & Average Decline \\ 
        \hline      
        GSWA(Self-Attn) & 1913 & 50.9 & 57.8 & 54.0 & 803 & 75.9  & 39.8 & 75.7 & 0.00\% \\
        GSWA(Cross-Attn)  & 1902 & 50.7 & 57.4 & 52.8 & 800 & 75.8 & 39.1 & 75.5 & 0.84\% \\
        GSWA(Cosine-Similarity)  & 1868 & 50.2 & 57.0 & 51.3 & 783 & 74.2 & 38.6 & 74.9 & 2.39\% \\
        \dag InternVL2-2B(w/o GSWA)  & 1862 & 49.9 & 56.6 & 50.1 & 785 & 73.2 & 37.5 & 74.3 & 3.49\% \\        
         \hline
    \end{tabular}}
    
\end{table*}

We consider the self-attention-based GSWA as the baseline for the weight allocation strategy and compare it with various alternative methods. Specifically, we perform ablation studies as follows: (1) replacing the multi-head self-attention in the GSWA module with multi-head cross-attention, (2) substituting the weight allocation scheme of the GSWA module with simple cosine similarity, and (3) directly removing the GSWA module. 

We initially replace the multi-head self-attention modules in the stacked transformer blocks of GSWA with multi-head cross-attention modules, which resulted in a decline in model performance across all benchmarks. For example, MME decreased from 1913 to 1902, and HRBench4K dropped from 54.0 to 52.8. This clearly demonstrates that the self-attention mechanism enables GSWA to allocate weights more appropriately than the cross-attention mechanism.

Next, we modify the weight allocation method in GSWA to calculate the cosine similarity between the  $<cls>$  token of each sub-image and the $<cls>$ token of the global  thumbnail. The calculated results are then normalized using the softmax function and used as the final allocated weights. This change lead to a significant decline in model performance across various benchmarks, with the MME score decreasing from 1907 to 1868 and OCRBench dropping from 800 to 783. In all benchmarks, the proportion of performance decline (2.39\%). These findings underscore the importance of designing a learnable, transformer-based adaptive sub-image weight allocation module.

Finally, we remove the GSWA module from the model and fine-tuned InternVL2-2B dierctly, resulting in a further decline in performance. Specifically, the TextVQA score decreases from 74.2 to 73.2, and the HallusionBench score drops from 38.6 to 37.5. Compared to the model utilizing our designed GSWA module, the average performance decreased by 3.49\%. This demonstrates the role of the GSWA module in enabling the model to focus on image regions with high information density, thereby enhancing performance across all visual understanding tasks.

\section{Limitation and Future Works}
While this work presents an effective approach to enhance the sub-image partitioning method for high-resolution image processing in LVLMs, several limitations must be addressed. The introduction of sub-image partitioning to LVLMs inherently results in the addition of extra visual tokens, which in turn leads to increased inference and training cost. Furthermore, the GSWA module also introduces additional computational overhead as an additional module, further contributing to the prolongation of inference and training time. Another limitation of our approach is that the GSWA module assigns weights to sub-images solely based on the provided high-resolution image itself, without considering the accompanying natural language input provided to the LVLM. This could result in sub-optimal weight distribution, as the relevance of certain sub-images may not fully align with the semantic context conveyed by the language input, potentially affecting the overall performance and accuracy of the model in corner cases. 

In future work, we aim to address the aforementioned limitations from two key aspects: (i) by integrating a text-guided weight allocator into the existing GSWA module, we seek to incorporate the provided natural language input when assigning weights to sub-images, thereby improving the alignment between the visual content and the semantic context conveyed by the language input; (ii) by introducing a token compression method to mitigate the increase in visual tokens resulting from sub-image partitioning, we plan to reduce the number of visual tokens, particularly from sub-images with lower weights, in order to ensure faster inference times while maintaining the model’s performance.

\section{Conclusion}
This paper aims to emulate the bottom-up visual attention mechanism observed in human vision to enhance the existing sub-image partitioning method for high-resolution image processing in LVLMs. We propose the GSWA module, which utilizes the self-attention mechanism to assign weights to sub-images derived from partitioning. We first conduct a comprehensive analysis of how sub-images from partitioning carry varying levels of information density. Building on this analysis, we introduce the GSWA module, which leverages semantic information from the global image to guide the allocation of weights to sub-images. This approach ensures the adaptive assignment of distinct weight values to sub-images based on their varying information densities, thereby avoiding the uniform processing of sub-images. By integrating the GSWA module into the InternVL2 framework, we develop SleighVL, which demonstrates exceptional performance across multiple task categories.

Despite certain limitations, including increased inference time, additional computational overhead, and the underutilization of textual input, our module introduces a novel approach that assigns weights to sub-images based on the mimicry of human visual behavior. This mechanism enables the model to focus more effectively on regions of the image with higher information density, therefore offering a fresh perspective for research in high-resolution image processing within LVLMs.

\bibliographystyle{unsrt}
\bibliography{references}

\end{document}